\documentclass[10pt,a4paper,onecolumn]{article}
\usepackage[utf8]{inputenc}
\usepackage[T1]{fontenc}
\usepackage{amsmath}
\usepackage{amsfonts}
\usepackage{amssymb}
\usepackage{graphicx}
\usepackage{caption}
\usepackage{subcaption}

\usepackage{lipsum}

\begin{document}

\title{ Study of the Fractal decomposition based  metaheuristic on low-dimensional Black-Box optimization problems}
\author{Arcadi Llanza, Nadiya Shvai, Amir Nakib\\
Universit\'e Paris Est Créteil, Laboratoire LISSI, Vitry Sur Seine, France\\
Vinci autoroutes, Cyclope.ai, Paris, France
}

\maketitle

\begin{abstract}
This paper analyzes the performance of the Fractal Decomposition Algorithm (FDA) metaheuristic applied to low-dimensional continuous optimization problems. This algorithm was originally developed specifically to deal efficiently with high-dimensional continuous optimization problems by building a fractal-based search tree with a branching factor linearly proportional to the number of dimensions. Here, we aim to answer the question of whether FDA could be equally effective for low-dimensional problems. For this purpose, we evaluate the performance of FDA on the Black Box Optimization Benchmark (BBOB) for dimensions 2, 3, 5, 10, 20, and 40. The experimental results show that overall the FDA in its current form does not perform well enough. Among different function groups, FDA shows its best performance on Misc. moderate and Weak structure functions.

\end{abstract}

\section{Introduction}
The general form of an optimization problem considered in this paper is defined by Eq. \ref{eq:opt}:
\begin{equation}
Min f(\vec{x}),  s.t. \vec{B_l} \le \vec{x} \le \vec{B_u}
\label{eq:opt}
\end{equation}

where $f(\vec{x})$ denotes the function to be minimize. It is assumed to be continuous. $\vec{x} = (x_1, x_2, ..., x_D)$ is the variable vector in $\mathbb{R}^D$. Here, $\vec{x}$ is a given parameter. Moreover, the function is constrained by $\vec{B_l} = (B_{l1}, B_{l2}, ..., B_{lD})$ as the lower boundary and $\vec{B_u} = (B_{u1}, B_{u2}, ..., B_{uD})$ as the upper boundary. 

The Fractal Decomposition Algorithm (FDA) is a deterministic metaheuristic method that has been shown to solve large scale (50 up to 1000 dimensions) continuous optimization problems with high-performance \cite{nakib2017deterministic}, \cite{nakib2019parallel}. This research aims to benchmark, for the first time, FDA in a low-dimensional (5 up to 40 dimensions) constrained continuous optimization problem such as Black Box Optimization Benchmark (BBOB) \cite{hansen2009real}.

In this paper, the Black Box optimization refers to the design and analysis of algorithms for problems where the structure of the objective function is unknown and unexploitable.
The rest of the paper is organized as follows. First, Section \ref{section:Related Work} reviews the related work. Then, Section \ref{section:Methodology} examines the foundations of the proposed method DFDA. Afterwards, Section \ref{section:Benchmark} describes the used benchmark. Section \ref{section:Results} presents the experiments and results. Finally, in Section \ref{section:Conclusions} the conclusion and further research directions are presented.

\section{Related work}
\label{section:Related Work}

To our knowledge, no other study has been previously done on FDA performance for low-dimensional continuous optimization problems. In the original FDA paper \cite{nakib2017deterministic}, the benchmark considered was SOCO2011. The problem dimension was set to a range of values from 50 to 1000. FDA ranked first for each considered.

An overview of state-of-the-art (SOTA) methods for the BBOB benchmark is provided. The methods are reported in order of their average performance from best to worst.
In the case of noiseless BBOB, generally, it is Evolutionary Algorithms (EAs) that perform better. Nevertheless, Local Searches (LS) and other hybrid methods are competitive as well. \textit{Hansen et al.} proposed in \cite{hansen2009benchmarking} a multistart BI-population CMA-ES with equal budgets for two interlaced restart strategies, one with increasing population size and one with varying small population sizes. In \cite{bosman2009amalgam}, \textit{Bosman et al.} introduced the Adapted Maximum-Likelihood Gaussian Model Iterated Density-Estimation Evolutionary Algorithm (AMaLGaM-IDEA). AMaLGaM-IDEA is a parameter-free algorithm with incremental model building (iaMaLGaM IDEA). MA-LS-Chain \cite{molina2009memetic} was proposed  by \textit{Molina et al.}. It uses a memetic algorithm with continuous local search. The Variable Neighbourhood Search (VNS) was suggested by \textit{Garcia et al.} in \cite{garcia2009continuous}. IPOP-SEP-CMA-ES \cite{ros2009benchmarking} is an algorithm with a multistart strategy with increasing population size introduced by \textit{Ros et al.}

The Age-Layered Population Structure (ALPS) Evolutionary Algorithm (EA) is a method presented by \textit{Hornby et al.} in \cite{hornby2009steady}. ALPS claims to avoid premature convergence than other EAs methods.

The Prototype Optimization with Evolved Improvement Steps (POEMS) was introduced by \textit{Kubalik et al.} in \cite{kubalik2009black}. POEMS is a stochastic local search-based algorithm.

The restarted estimation of distribution algorithm (EDA) with Cauchy distribution  (Cauchy-EDA) probabilistic model was suggested by \textit{Povsik et al.} in \cite{povsik2009bbob}. Cauchy-EDA claims to be usable for many fitness landscapes. On the contrary, EDA with Gaussian distribution tends to converge prematurely.

The Differential Ant-Stigmergy Algorithm (DASA) was presented in \cite{korovsec2009stigmergy} by \textit{Korovsec et al.} DASA is a stigmergy-based algorithm for solving optimization problems with continuous variables.

\textit{Hansen et al.} analysed the Nelder-Mead downhill simplex method  \cite{hansen2009benchmarking}. Nelder-Mead is a multistart strategy applied on local and global levels. On the one hand, at the local level, ten restarts are conducted with a small number of iterations and reshaped simplex. On other hand, at the global level, independent restarts are launched until 105D function evaluations are exceeded.

\section{The Fractal decomposition based Algorithm}
\label{section:Methodology}

\begin{figure}[h]
	\centering
	\includegraphics[height=5cm]{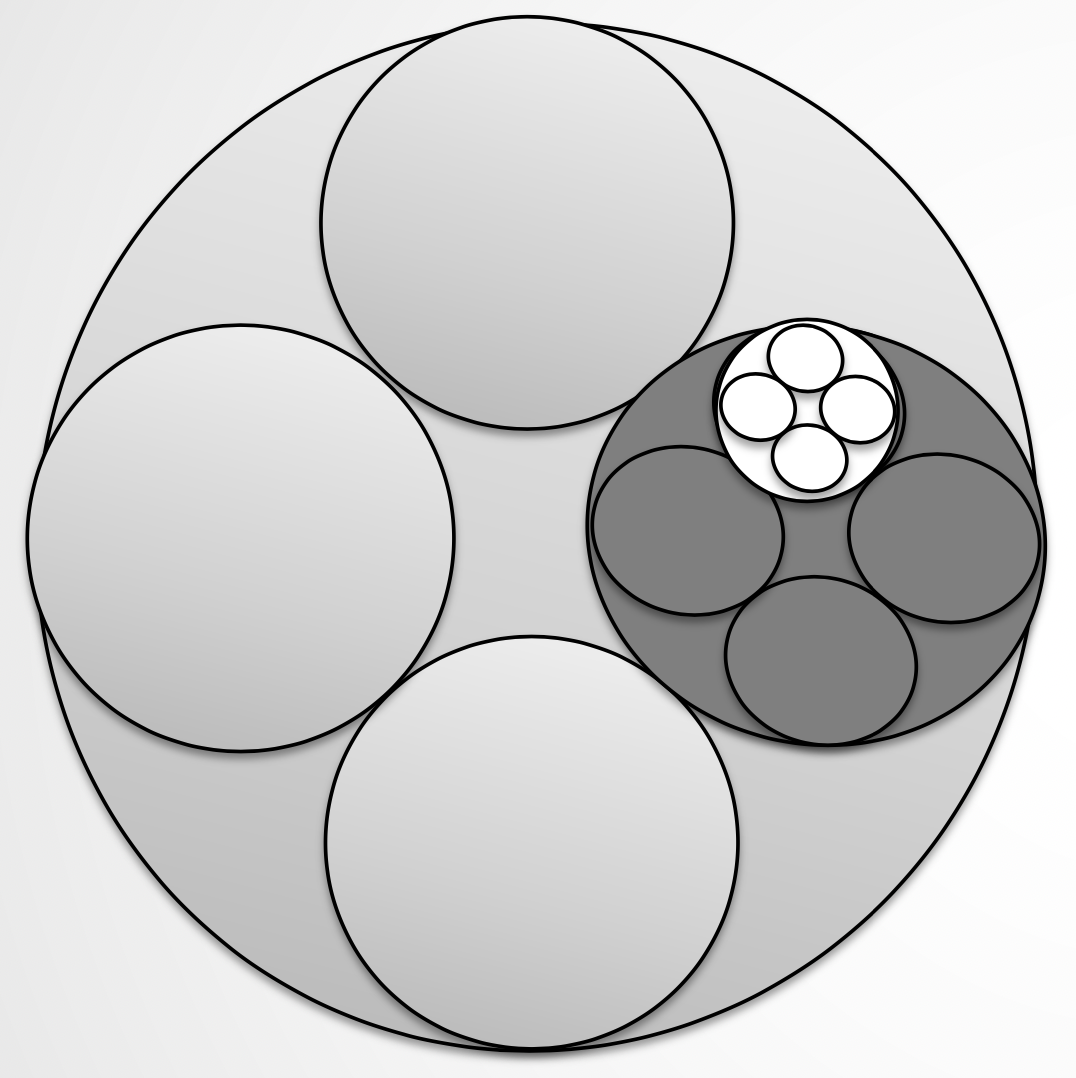}
	\hspace{1cm}
	\includegraphics[height=5cm]{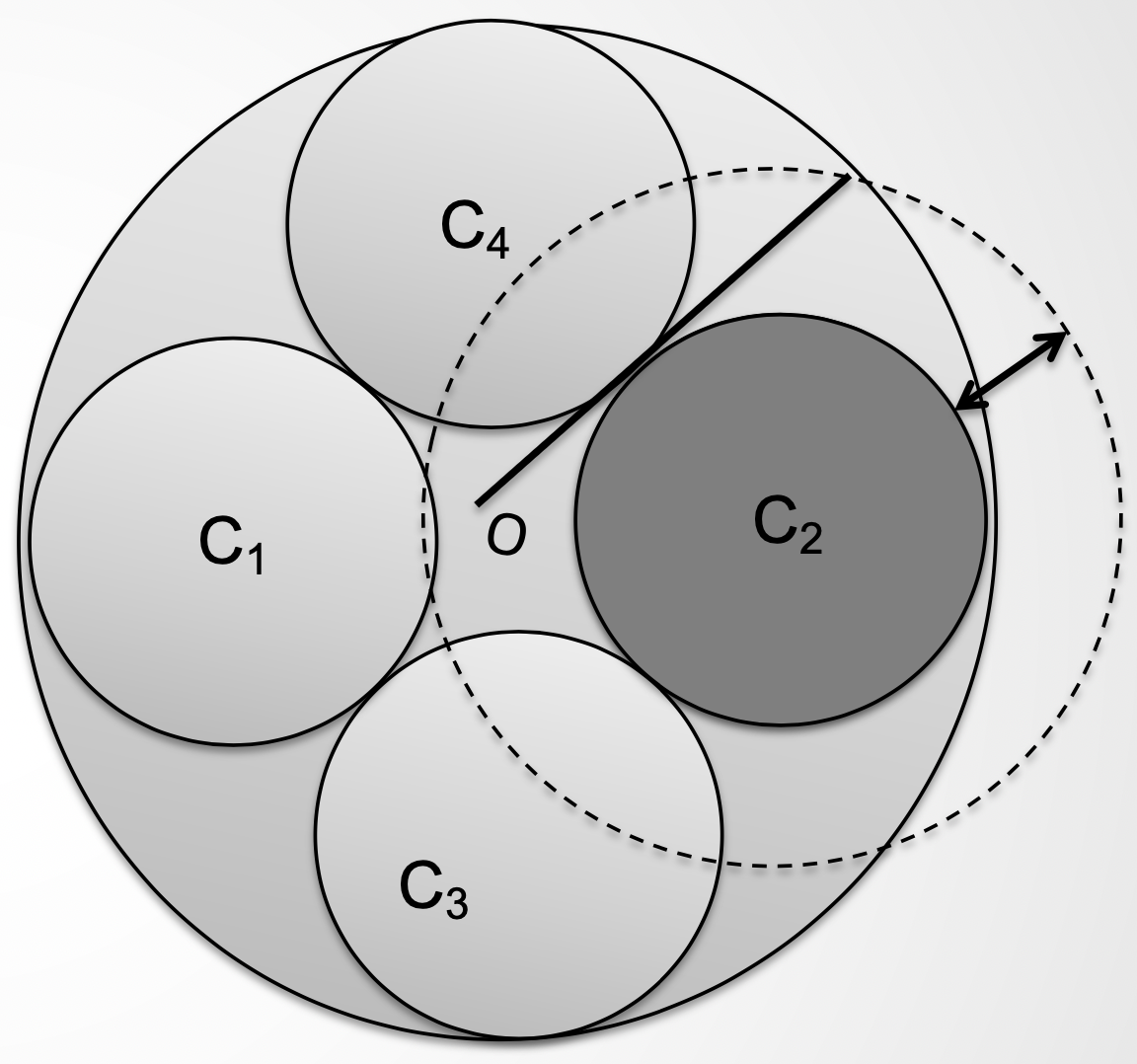}
	\caption{Hypersphere fractal decomposition at 4 levels of depth \cite{nakib2017deterministic}.}
	\label{fig:FDA_decomposition}
\end{figure}

Nakib~\textit{et al.} introduced a fractal decomposition \cite{nakib2017deterministic} based on hyperspheres to solve high dimensional continuous optimization problems with low complexity.
FDA \cite{nakib2017deterministic} uses two components to find the optima
in the landscape: the First component, called fractal decomposition, is used as an exploration technique. Then, the second component, called Intensive Local Search (ILS), is used as the exploitation technique to search in the local regions previously identified as promising regions. The basic pattern used in the fractal decomposition is a hypersphere because when scaled into a high dimensional space, its computational complexity is low. FDA  covers the space with few hyperspheres allowing it to obtain a good performance in its exploratory phase. An inflation procedure is applied to the hyperspheres (see Fig. \ref{fig:FDA_decomposition}) to ensure that all space is covered.

The following subsection is dedicated to the different FDA components.

\subsection{Exploration component}
At this phase, promising regions are searched for by conveniently subdividing the search space into smaller regions that might contain a good solution. The partition of space is modeled after a form of a hypersphere. This shape is a suitable representation that allows FDA to be extremely competitive in large-scale spaces.

Given a center $C_k$ of a hypersphere with radius $r$ the centers of its decomposition can be obtained as in Eq. \ref{eq:hyperspheres_equation}:
\begin{equation}
C^{i}_{k+1} = C_{k} + (-1)^i \cdot ((r - r^\prime) e_{[\frac{i + 1}{2}]})
\label{eq:hyperspheres_equation}
\end{equation}

Then, the quality of each generated hypersphere is evaluated based on two points $\overrightarrow{s_1}$ and $\overrightarrow{s_2}$ originated based on Eq. \ref{eq:FDA_quality_assessment}.

\begin{eqnarray}
\overrightarrow{s_{1}}=\overrightarrow{C}^{l}+\alpha \frac{r_{l}}{\sqrt{D}} \times \overrightarrow{e}_{d}, \: \text { for } \: d=1,2, \ldots, D \nonumber
\\
\overrightarrow{s_{2}}=\overrightarrow{C}^{l}-\alpha \frac{r_{l}}{\sqrt{D}} \times \overrightarrow{e}_{d}, \: \text { for } \: d=1,2, \ldots, D
\label{eq:FDA_quality_assessment}
\end{eqnarray}

Subsequently, for the aforementioned positions ($\overrightarrow{s_1}$, $\overrightarrow{s_2}$) and the center of the hypersphere $\overrightarrow{C_l}$, their fitnesses, $f_1$, $f_2$, and $f_c$, respectively, are evaluated. Furthermore, the distances to the best position found so far (BSF) are also computed via the $L^2-norm$.

During this process, it is important to point out that the best solutions and their coordinates are saved.
At this point, the slope $g_1$, $g_2$, and $g_c$ is calculated on the three positions' fitness ($f_1$, $f_2$, and $f_c$) based on the $L_2 \: norm$ distance to BSF as in Eq. \ref{eq:ratio_computation}.

\begin{equation}
g_{1}=\frac{f\left(\overrightarrow{s_{1}}\right)}{\left\|\overrightarrow{s_{1}}-B S F\right\|}, g_{2}=\frac{f\left(\overrightarrow{s_{2}}\right)}{\left\|\overrightarrow{s_{2}}-B S F\right\|} \, \, \text { and } \,\, g_{c}=\frac{f\left(\overrightarrow{C^{l}}\right)}{\left\|\overrightarrow{C^{l}}-B S F\right\|}
\label{eq:ratio_computation}
\end{equation}

The quality of the hypersphere will be defined by the highest ratio among $g_1$, $g_2$, and $g_c$, denoted by $q$ as in Eq. \ref{eq:hypersphere_quality}:

\begin{equation}
q=\max \left\{g_{1}, g_{2}, g_{c}\right\}
\label{eq:hypersphere_quality}
\end{equation}

As an important remark, each level of hyperspheres that has not yet been decomposed, is stored in a list and sorted by its quality score. Therefore, when all the hyperspheres in a level have been explored, the next level of hyperspheres unlocks until the stopping criterion is reached. By default the stopping criterion is defined on one of the following conditions:
\begin{itemize}
	\item The maximum number of evaluations given by the benchmark is reached.
	\item The maximum decomposition level $k$ is reached.
\end{itemize}

\subsection{Exploitation component}

In the intensification phase, ILS is used for its simplicity and efficiency to find a local/global optimum at the end of its execution.

This technique uses two candidate solutions ($\vec{x}^{s_{1}}$ and $\vec{x}^{s_{2}}$) that are evaluated per dimension. Each candidate is located one step size ($\omega$) in the opposite directions \textit{w.r.t.} the current solution $\vec{x}^{s}$ based on Eq. \ref{eq:candidate_location}:

\begin{eqnarray}
&\vec{x}^{s_{1}}=\vec{x}^{s}+\omega \times \vec{e}_{i}  \nonumber
\\
&\vec{x}^{s_{2}}=\vec{x}^{s}-\omega \times \vec{e}_{i}
\label{eq:candidate_location}
\end{eqnarray}

where $\vec{e}_i$ is the unit vector in which the i-th element is set to 1 and the other elements to 0. 

Then the best solution among $\vec{x}^{s}$, $\vec{x}^{s_{1}}$, and $\vec{x}^{s_{2}}$ is selected to be the next current solution $\vec{x}^{s}$. The factor $\omega$ is halved whenever ILS cannot find a better solution in any of the dimensions until a stopping criterium is reached. This can happen in any of the following cases:

\begin{itemize}
	\item The maximum number of evaluations given by the benchmark is reached.
	\item $\omega$ keeps decreasing until a given value $\omega_{min}$ that denotes the tolerance or the minimum precision needed by the benchmark.
\end{itemize}

\section{Benchmark}
\label{section:Benchmark}

BBOB is a continuous optimization problem with mixed-integer domains which consists of a set of 6 suits (\textit{bbob}, \textit{bbob-noisy}, \textit{bbob-biobj}, \textit{bbob-largescale}, \textit{bbob-mixint}, \textit{bbob-biobj-mixint}). In this study, we analyze the \textit{bbob} suit that consists of a set of functions (f1 - f24) divided into five groups (Separable, Misc. moderate, Ill-conditioned, Multi-modal, and Weak structure) which are scalable to the dimension. The functions are defined within the hypercube $[-5, 5]^D$, where $D$ is the dimensionality of the search space. Furthermore, each function has 15 instances to ensure results are statistically significant when reported. Generally, the difficulty in BBOB increases from the first to the last group. Groups are used to aggregate the obtained results, into more meaningful reports on the performance of functions with particular characteristics.

To compare real-parameter global optimizers, we used the COmparing Continuous Optimizers (COCO) framework \cite{hansen2021coco}. COCO provides benchmark function testbeds, easy-to-parallelize experimentation templates, and tools for processing and visualization tools for data generated by one or more optimizers.

\subsection{Performance metrics}
Average execution time ($\mathrm{aRT}$) was introduced in \cite{price1997differential} under the name as ENES and afterwards referred to in \cite{auger2005performance} and \cite{hansen2009benchmarking} as success performance and ERT correspondingly. This metric estimates the expected execution time of the restart algorithm. Typically, the average over all the trials is taken by varying only the reference instantiation parameters $\theta_i$.

The execution time of the restart algorithm is given in Eq. \ref{eq:RT}. Here, we imply that the instance of the optimization problem  $p = (n, f_\theta, \theta_i)$ is given by triple of search space dimension, function, and instantiation parameters. The subscripts \textit{us} and \textit{s} denote unsuccessful and successful trials, $\Delta I$ is the precision, and $J \sim \mathrm{BN}(1, 1-p_s)$ is a random variable with negative binomial distribution that models the number of unsuccessful runs until a success is observed given $p_\mathrm{s} > 0$ the success probability of the algorithm.

\begin{equation}
\mathbf{RT}(n, f_\theta, \Delta I) = \sum_{j=1}^{J} \mathrm{RT}^{\rm us}_j(n,f_\theta,\Delta I) + \mathrm{RT}^{\rm s}(n,f_\theta,\Delta I)
\label{eq:RT}
\end{equation}

Given a dataset that succeeds at least once ($n_\mathrm{s}\ge1$) with runtimes $\mathrm{RT}^{\rm s}_i$, and $n_\mathrm{us}$ unsuccessful runs with $\mathrm{RT}^{\rm us}_j$ evaluations, the average runtime is expressed as in Eq. \ref{eq:aRT}:

\begin{eqnarray}
\mathrm{aRT}
& = &
\frac{1}{n_\mathrm{s}} \sum_i \mathrm{RT}^{\rm s}_i +
\frac{1-p_{\mathrm{s}}}{p_{\mathrm{s}}}\,
\frac{1}{n_\mathrm{us}} \sum_j \mathrm{RT}^{\rm us}_j \nonumber
\\
& = &
\frac{\sum_i \mathrm{RT}^{\rm s}_i + \sum_j \mathrm{RT}^{\rm us}_j }{n_\mathrm{s}} \nonumber
\\
& = &
\frac{\#\mathrm{FEs}}{n_\mathrm{s}}, \label{eq:aRT}
\end{eqnarray}

where $\# \mathrm{FEs}$ denotes the total number of function evaluations made in all trials before reaching the target precision.

\section{Experiments and discussion}
\label{section:Results}

\begin{figure}[]
	\centering
	\includegraphics[width=\textwidth]{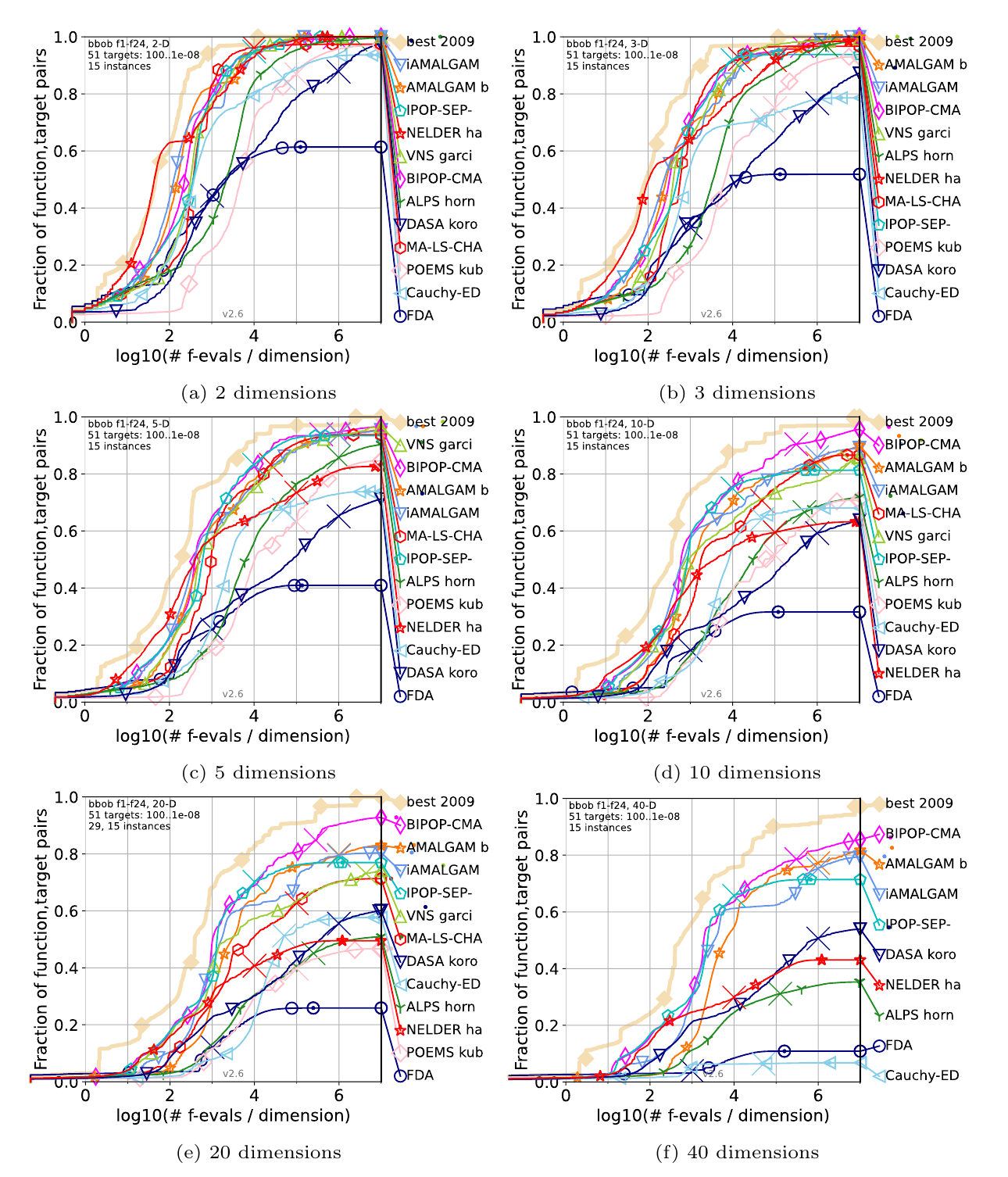}
	\\
	\caption{Comparison of 12 SOTA methods per dimensions 2, 3, 5, 10, 20, and 40. The x-axis represents a particular budget in time (evaluations). The y-axis represents the performance of the 24 bbob functions that the given method has managed to solve (the higher the value the better). Each function is represented by 15 instances.}
	\label{fig:SOTA}
\end{figure}

In the following subsections, the results of the 24 BBOB functions are analyzed. FDA has been benchmarked on the dimensions $D = 2, 3, 5, 10, 20, 40.$ The maximum number of function evaluations (maximum budget) is chosen as $1000\times D$. Experimental results show that FDA performs best on the separable functions. However, it works also quite well on Misc. moderate functions with lower dimensions. On the other hand, the algorithm fails to solve optimization problems with functions based on a multi-modal structure. The summary of FDA performance comparing to benchmark SOTA is provided in Figure \ref{fig:SOTA}.


The rest of the experiments section is organized as follows. First, function results are broken down by dimension and function group to better understand FDA performance. Then, an aggregated graph is offered to summarize the previously mentioned information and expand the view of the dimensions. Afterward, target precision details based on FDA evaluation consumption are supplied. Finally, a comparison with other methods mentioned in previous sections is provided.

\subsection{Runtime distributions (ECDFs) summary and function groups}

\begin{figure}[h]
	\centering
	\includegraphics[width=\textwidth]{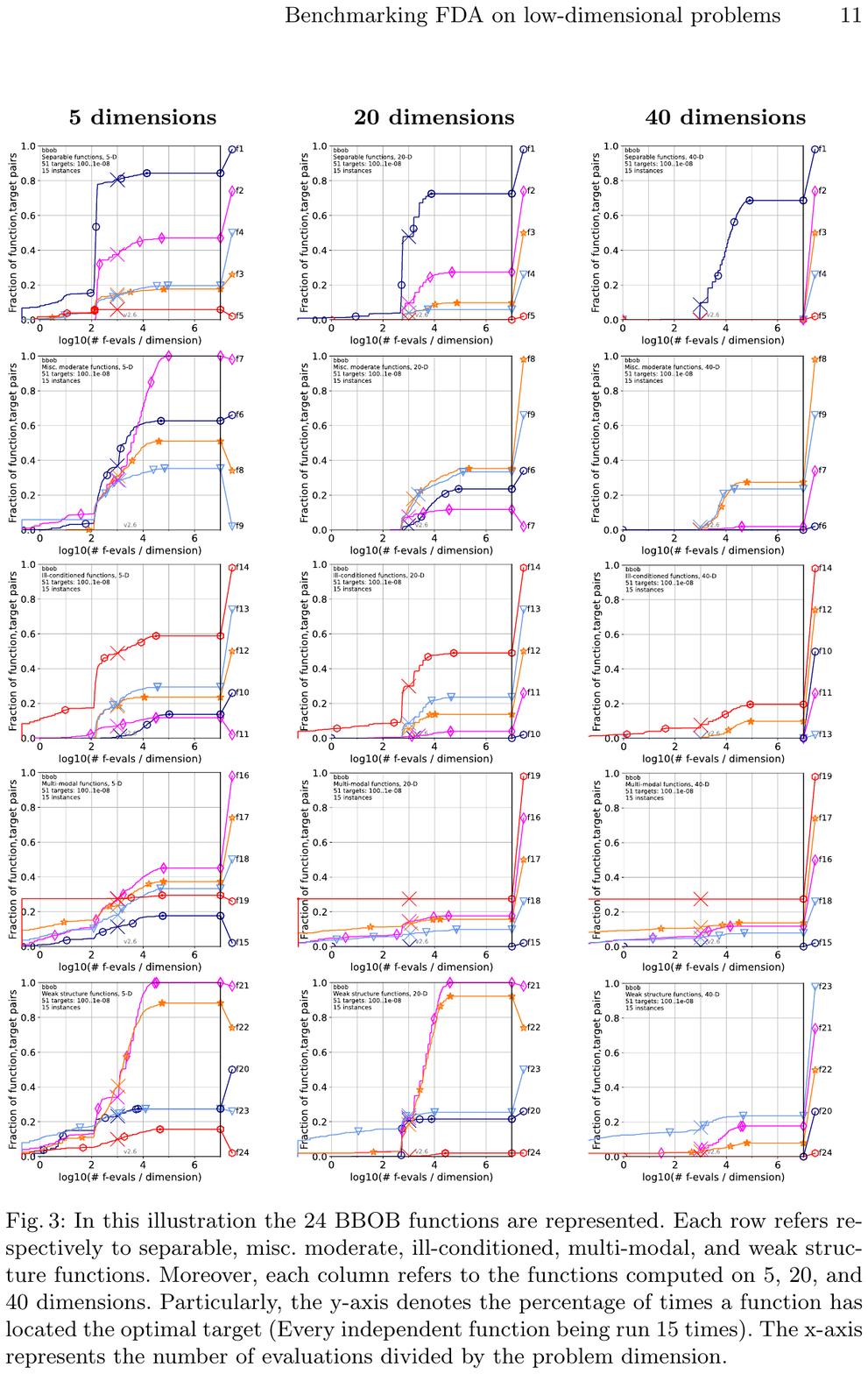}
	\\
	\caption{In this illustration the 24 BBOB functions are represented. Each row refers respectively to separable, misc. moderate, ill-conditioned, multi-modal, and weak structure functions. Moreover, each column refers to the functions computed on 5, 20, and 40 dimensions. Particularly, the y-axis denotes the percentage of times a function has located the optimal target (Every independent function being run 15 times). The x-axis represents the number of evaluations divided by the problem dimension.}
	\label{fig:Runtime_distributions}
\end{figure}

FDA performance by function groups can be observed in Figure \ref{fig:Runtime_distributions}. Each column represents a different dimension complexity. FDA has been tested in dimensions 2, 3, 5, 10, 20, and 40. However, only dimensions 5, 10, and 40 have been chosen to be shown in this graph. Each column depicts the group of functions available in BBOB (separable, misc. moderate, ill-conditioned, multi-modal, and weak structure functions). Overall, FDA does not perform well on the benchmark. In particular, often it does not succeed finding the optima. FDA performs better in the Misc. moderate and Weak structure functions. Nevertheless, it does not reach a minimum performance standard in this low-dimensional problem.

\begin{figure}[h]
	\centering
\includegraphics[width=\textwidth]{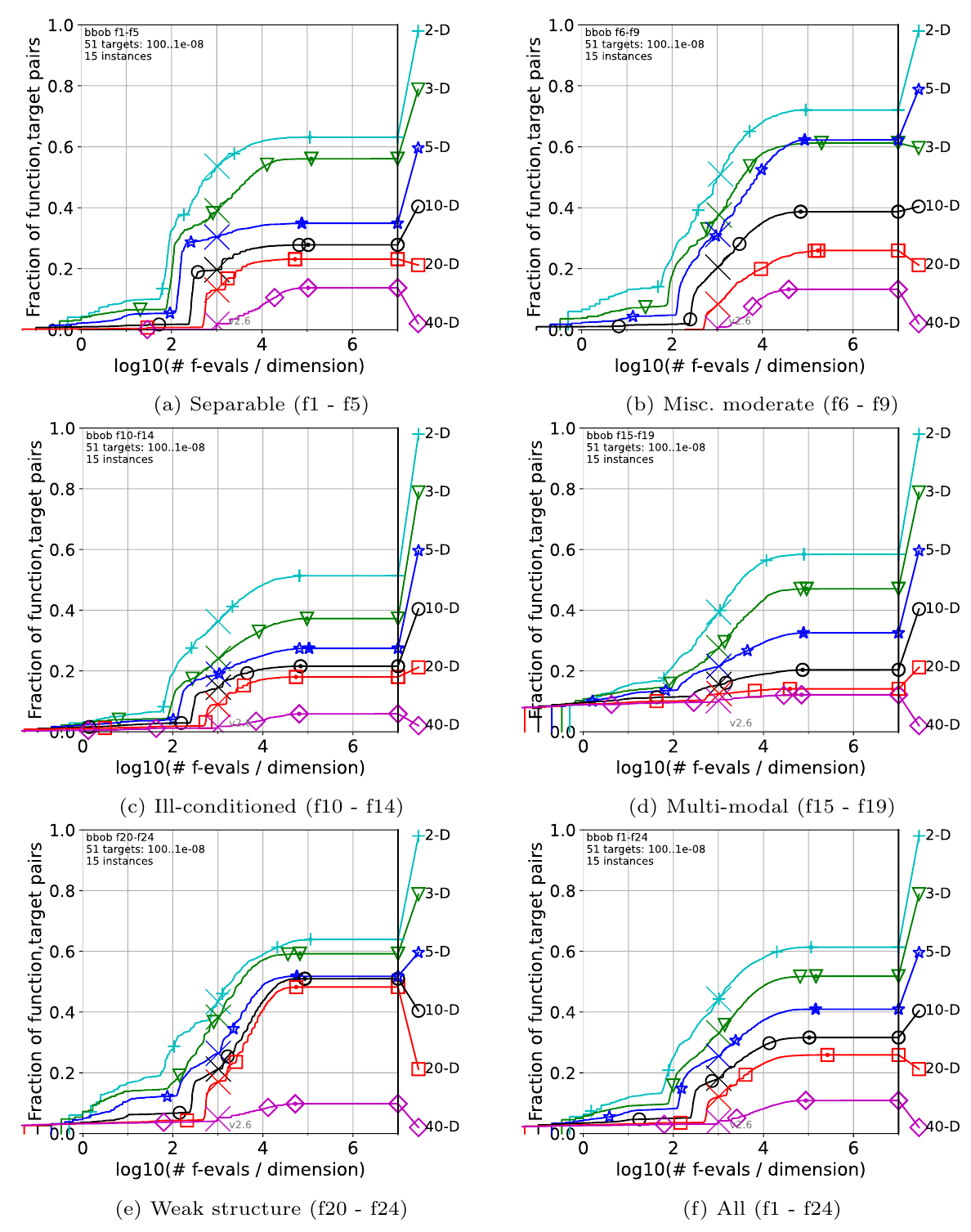}
	
	\caption{Summary of function groups per dimension. The y-axis denotes the percentage of times a function has located the optimal target (Every independent function being run 15 times). The x-axis represents the number of evaluations divided by the problem dimension.}
	\label{fig:summary_Runtime_distributions}
\end{figure}

In Figure \ref{fig:summary_Runtime_distributions} the aforementioned summary including all the dimensions where FDA was benchmarked is provided. Each graph compares a set of functions per dimension. The higher the dimension the more complex the problem becomes. As it can be observed, in many cases FDA does not reach the global optimum.

\subsection{Scaling of runtime with problem dimension}

In Figure \ref{fig:Scaling_of_runtime} the expected runtime per target function precision \textit{w.r.t.} dimension is presented. The values obtained are plotted in a logarithmic scale. The symbol $+$ represents the median run time of successful runs to complete the hardest goal that was completed at least once (but not always). The symbol $\times$ characterizes the maximum number of function evaluations in a trial. The FDA attempted to adjust only once per instance function due to its deterministic nature.

\begin{figure}[h]
	\centering
	\includegraphics[width=\textwidth]{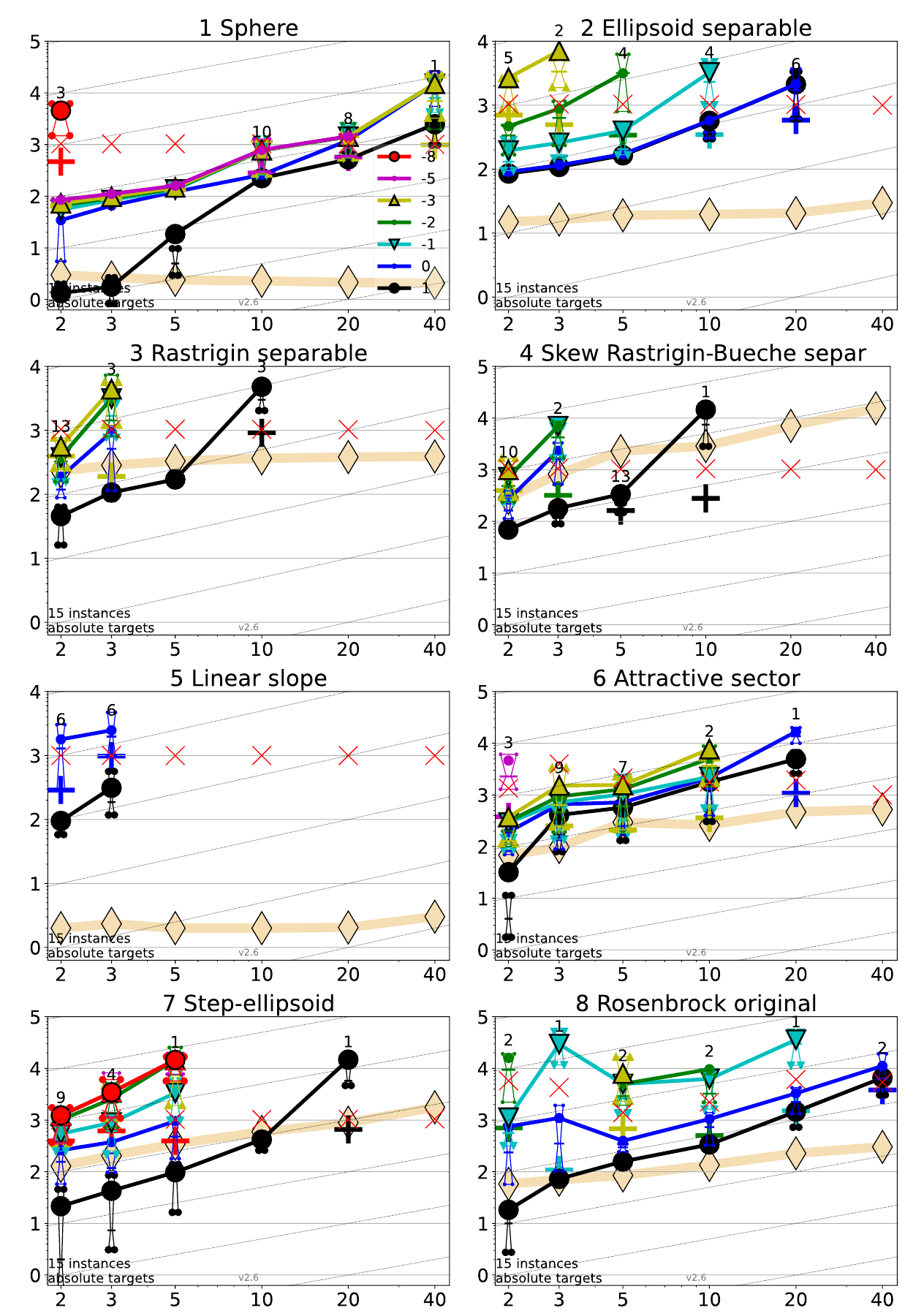}
	
\end{figure}

\begin{figure}[h]\ContinuedFloat
\includegraphics[width=\textwidth]{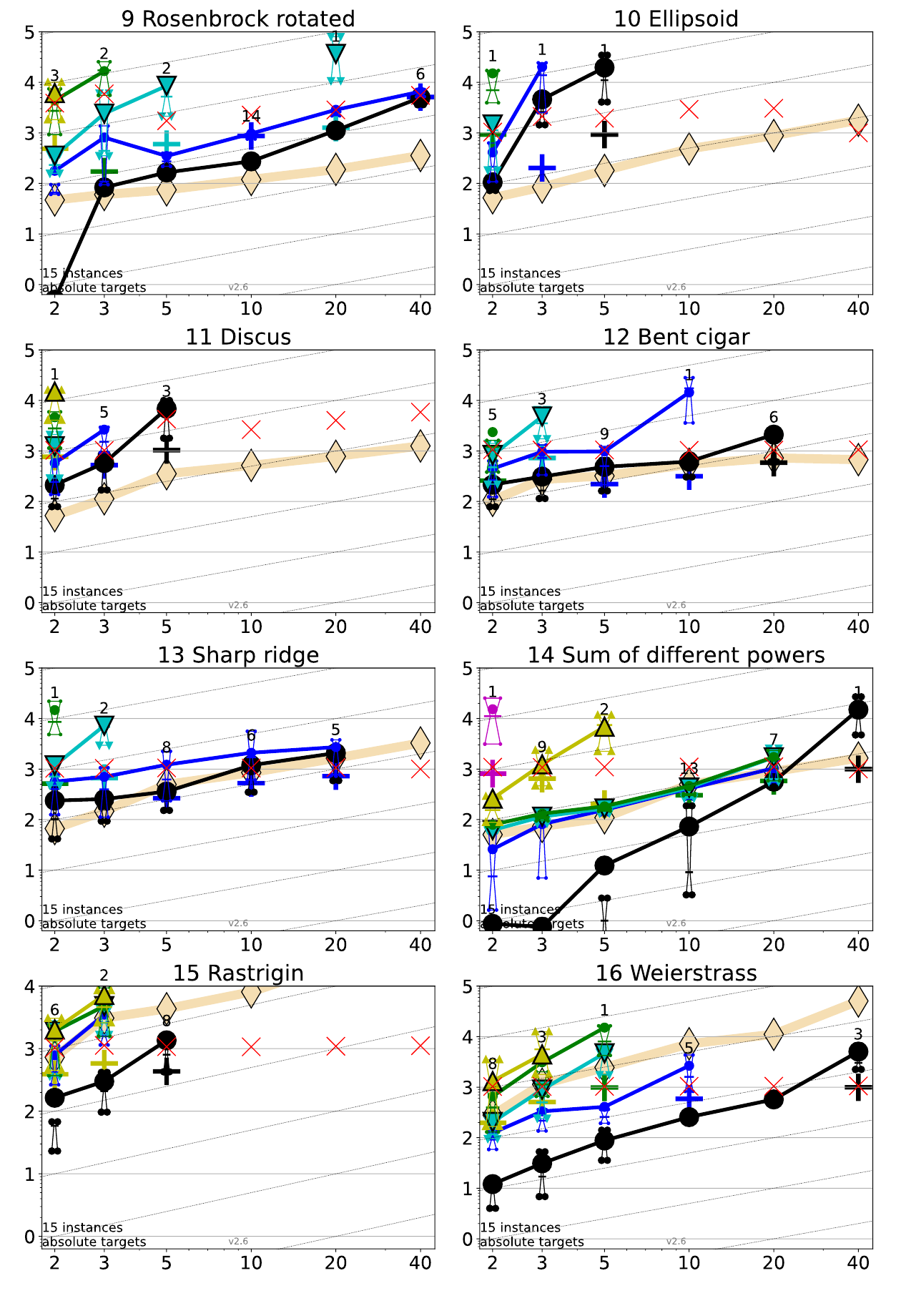}
\end{figure}

\begin{figure}[h]\ContinuedFloat
\includegraphics[width=\textwidth]{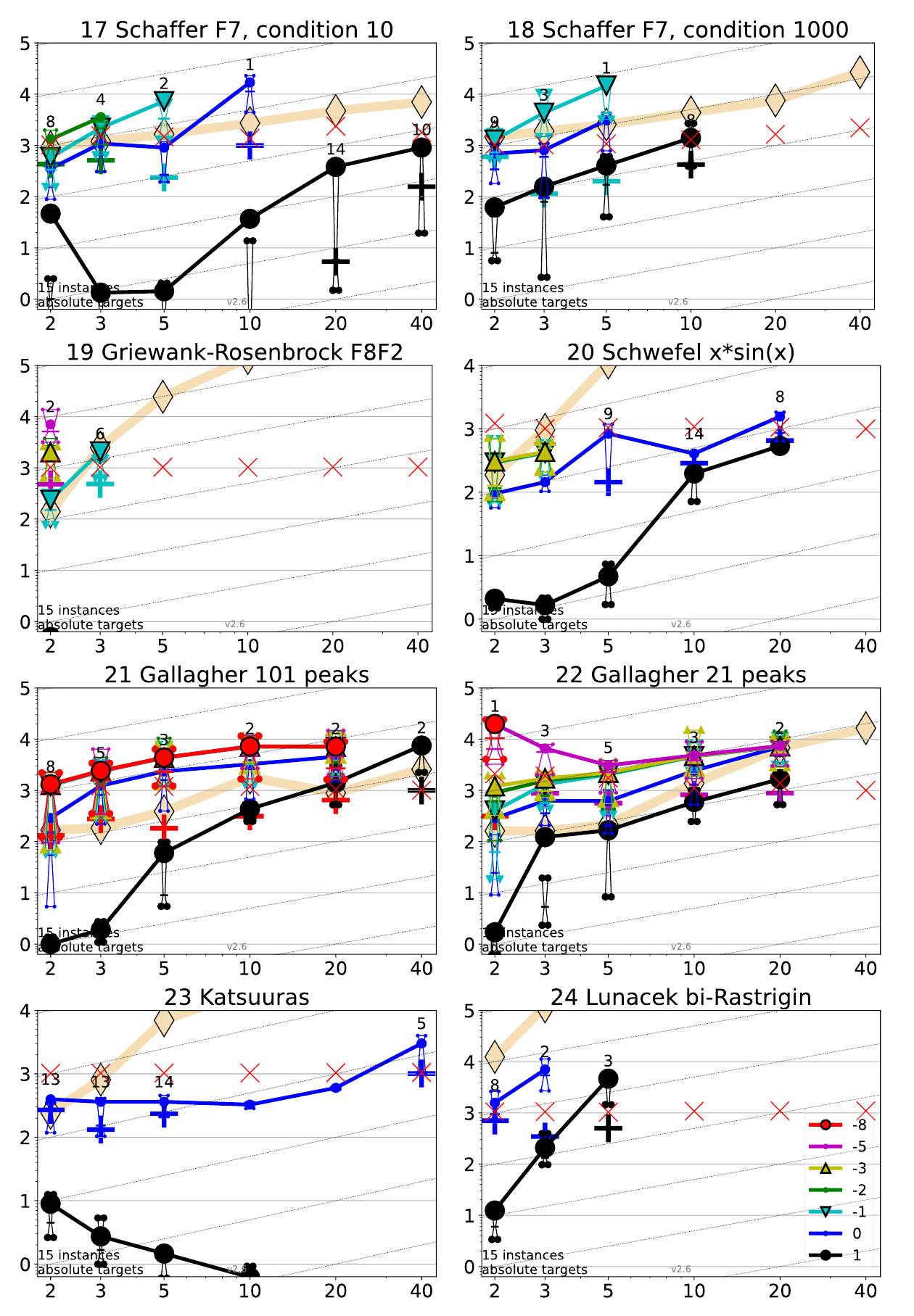}
	\caption{In this figure the 24 BBOB functions are displayed at different target precision. The diamond shape in the graphs presents the best method in 2009 at the most complex target (red line). The y-axis denotes the expected runtime (ERT; the lower the better). The x-axis shows the problem dimension. Finally, as a legend is used the ERT for fixed target precision values of 10k.}
	\label{fig:Scaling_of_runtime}
\end{figure}

\subsection{Discussion}
We find that the FDA is not adapting well to the current problem. Nonetheless, the FDA has a promising start in exploring the search space with its fractal-based technique that subdivides space into promising smaller regions. On the contrary, ILS turns out to be too slow in the intensification phase harming the good initial performance. In particular, the main disadvantage of the FDA intensification step is not being able to abandon an unpromising solution trajectory.

\bibliographystyle{plain}
\bibliography{MicArxiv}

\end{document}